\documentclass[10pt,letterpaper]{article}

\usepackage{cogsci}
\usepackage{graphicx}
\cogscifinalcopy % Uncomment this line for the final submission 

\usepackage{pslatex}
\usepackage{apacite}
\usepackage{float} % Roger Levy added this and changed figure/table
\usepackage{boldline}
\usepackage{natbib}

\renewcommand\footnotemark{}

\title{A Relational Inductive Bias for Dimensional Abstraction in Neural Networks}
 
\author{%
Declan Campbell \textsuperscript{1}  \textbf{Jonathan D. Cohen \textsuperscript{1, 2}} \\[1ex]
1 Princeton Neuroscience Institute \\
2 Princeton University Department of Psychology \\
}

\begin{document}

\maketitle

\begin{abstract} 
The human cognitive system exhibits remarkable flexibility and generalization capabilities, partly due to its ability to form low-dimensional, compositional representations of the environment. In contrast, standard neural network architectures often struggle with abstract reasoning tasks, overfitting, and requiring extensive data for training. This paper investigates the impact of the relational bottleneck—a mechanism that focuses processing on relations among inputs—on the learning of factorized representations conducive to compositional coding and the attendant flexibility of processing. We demonstrate that such a bottleneck not only improves generalization and learning efficiency, but also aligns network performance with human-like behavioral biases. Networks trained with the relational bottleneck developed orthogonal representations of feature dimensions latent in the dataset, reflecting the factorized structure thought to underlie human cognitive flexibility. Moreover, the relational network mimics human biases towards regularity without pre-specified symbolic primitives, suggesting that the bottleneck fosters the emergence of abstract representations that confer flexibility akin to symbols.

\end{abstract}

\section{Introduction}

The flexibility and generativity of human cognition has been the focus of intense research in cognitive science and machine learning.  \citet{fodor1988connectionism} famously proposed that this depends on the property of compositionality supported by symbolic processing which, they argued, was not supported by neural network architectures. In contrast, a strong focus and much of the success of early neural network architectures \citep{ mcclelland1986appeal, rumelhart1986learning}, as well as modern forms of deep learning \citep{lecun2015deep}, has been on the ability to implement sophisticated forms of statistical learning. In addition to achieving powerful forms of function approximation, these architectures have been shown to achieve capabilities traditionally associated with symbolic processing \citep{rogers2004semantic}, including the ability of transformer-based, large-language models to solve challenging analogical reasoning tasks \citep{webb2023emergent}. Nevertheless: a) these models fail on other forms of abstract reasoning tasks on which humans succeed \citep{mitchell2023comparing};  b) they require amounts of data and training that are far in excess of what people require to achieve comparable performance \citep{mnih2015human}; and c) at best, it is unclear whether they make use of the same kinds of low dimensional, highly abstract, compositional (and, in the limit, genuinely symbolic) forms of representation that Fodor and Pylyshyn had in mind. 

Interestingly, it has recently been shown that introducing simple inductive biases to standard neural network architectures can promote the efficient learning of and flexible use of abstract representations and, in some cases, the emergence of genuinely symbolic forms of processing \citep{webb2020emergent, kerg2022neural, altabaa2023abstractors}. These inductive biases implement a form of relational bottleneck: a processing component that restricts the flow of information to relations among the inputs \citep{webb2023relational}, upon which further processing is based. This isolation between processing of the data and processing that is restricted to only relational information inherent in the data forces the latter to learn abstract representations, and functions that use these, that can be generalized to new, never seen inputs that exhibit the same relations. 

One corollary of the idea that a relational bottleneck can induce the learning of abstract representations is that this should be facilitated by representation of data in a compositional form — that is, factorized in a form that the relational structure is most apparent. The learning of such factorized representations are a requisite for compositional coding. One example of such factorization that has been an important goal of research on deep learning \citep{higgins2016beta, kim2018disentangling} is the learning of dimensional structure that may exist in the data (e.g., the color, size, and brightness of visual images). Here, we explore the possibility that the relational bottleneck can play an important role in promoting the learning of such representations. Specifically, we test the hypothesis that, insofar as factorized representations of distinct, task-relevant feature dimensions make learning relations easier, then imbuing a network with a relational bottleneck may put pressure on the network to learn such representations, improving generalization performance on relational tasks, and better aligning network performance with human behavioral biases.

To test this hypothesis, we focused on the learning of similarity relations using the simplest form of a relational bottleneck, which can be implemented in a contrastive network. Specifically, we implemented two standard multilayer perceptron (MLP) pathways with shared encoder weights (as input streams for the two items to be compared), that converged on a layer in which the distance (e.g., cosine similarity) was computed between pairs of embeddings. This was used to evaluate the similarity between the embeddings in each input stream, that was taken as the response.  The network was trained to produce the correct response (i.e., how similar the two inputs were) for each pair of stimuli. This implementation of the relational bottleneck is comparable to the CoRelNet architecture described by \cite{kerg2022neural}. We compared the performance of this network, and the representations it learned, with a standard feedforward neural network that lacked a relational bottleneck (i.e., the direct similarity computation).

We carried out two sets of simulations to evaluate these networks. In the first, we evaluated the extent to which they developed orthogonal (factorized) representations of distinct feature dimensions in the embedding layers, using a training dataset comprised of features that varied along two orthogonal dimensions. In the second set of simulations, we extended the approach to a richer feature space of geometric forms previously generated using symbolic primitives, and tested the extent to which the networks captured the difference between performance of humans and non-humans that has previously been explained using models explicitly imbued with symbolic primitives \citep{dehaene2022symbols, sable2021sensitivity, sable2022language}.

\section{Relational Bottleneck and Dimensional Representations}

Our initial investigation focused on a simple similarity judgment task, to evaluate the extent to which a relational bottleneck induced the learning of factorized representations of two orthogonal feature dimensions in its embeddings. In contrast to existing unsupervised methods for learning factorized representations \citep{higgins2016beta, kim2018disentangling}, we used a contrastive loss function that measures the relation (distance) between pairs of inputs within the metric space defined by the dataset’s latent features. We show that this approach encourages the model to learn “factorized” representations of the task features. We then investigate the degree to which this factorization contributes to learning and generalization. 

\subsection{Methods}

\subsubsection{Networks}
We compared two simple forms of feedforward neural networks that differed only in how they computed stimulus similarity (Fig. 1). For the relational model, we implemented a simple form of the relational bottleneck (comparable to \cite{kerg2022neural}; \cite{altabaa2023abstractors}), in which similarity was computed directly as the Euclidean distance between pairs of embeddings learned by the encoding layers; this was then used to determine the response (red pathway in Fig. 1). We compared this to a standard feedforward network, without any explicit similarity computation, in which the encoding layer projected to a multilayer perceptron (MLP) that generated the response (blue pathway in Fig. 1).

\begin{figure}[h]
\centering 
\includegraphics[width=\linewidth]{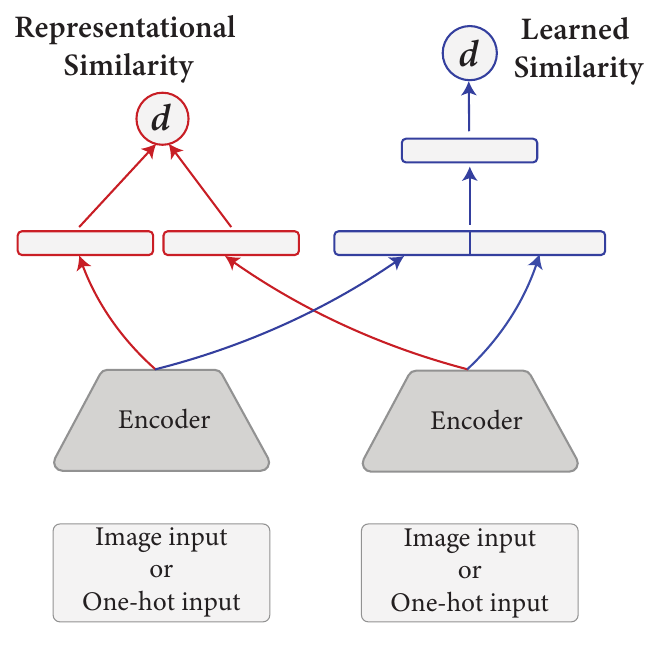}
\caption{\textbf{Network Architecture} Feedforward networks used to perform identity and similarity judgement tasks over two input stimuli (see text).}
\label{fig:model_architecture}
\end{figure}

\subsubsection{Task and training data} The task involved making similarity judgements over pairs of inputs that varied parametrically along two orthogonal dimensions. The inputs were pairs of grayscale images depicting geometric forms in which size and luminosity varied parametrically and independently across pairs (in the Appendix, we report similar results for the case in which features of the inputs were encoded as one-hots, rather than varying parametrically).

\subsection{Results}
First we examined both the rate at which the two networks learned to perform the task, and at which out of distribution (OOD) generalization improved — that is how quickly they learned to generalize performance to stimuli that were not only held out of the training corpus but that had feature values outside the range of those used in the training corpus. We found that that the relational network both learned to master the training data and exhibit OOD generalization substantially faster than the standard network (Fig. 2a). Next, we examined the structure of the representations learned in the encoding layers of the networks using PCA. Fig. 2b shows that the relational network learned orthogonal representations of the two feature dimensions, but that this was not the case for the standard network. These findings clearly indicate that imposing a simple form of the relational bottleneck in a standard feedforward network not only improves sample efficiency and the rate at which OOD generalization is achieved, but also imposes factorial structure on the representations learned immediately prior to the bottleneck — a form of representation that is fundamental to compositional coding and that is known to be a feature of, and fundamental to the flexibility of human cognitive function. In the next set of simulations, we more directly examined the extent to which the relational bottleneck reproduces empirical observations about biases in human similarity judgments.

\begin{figure*}[h]
\centering 
\includegraphics[width=\linewidth]{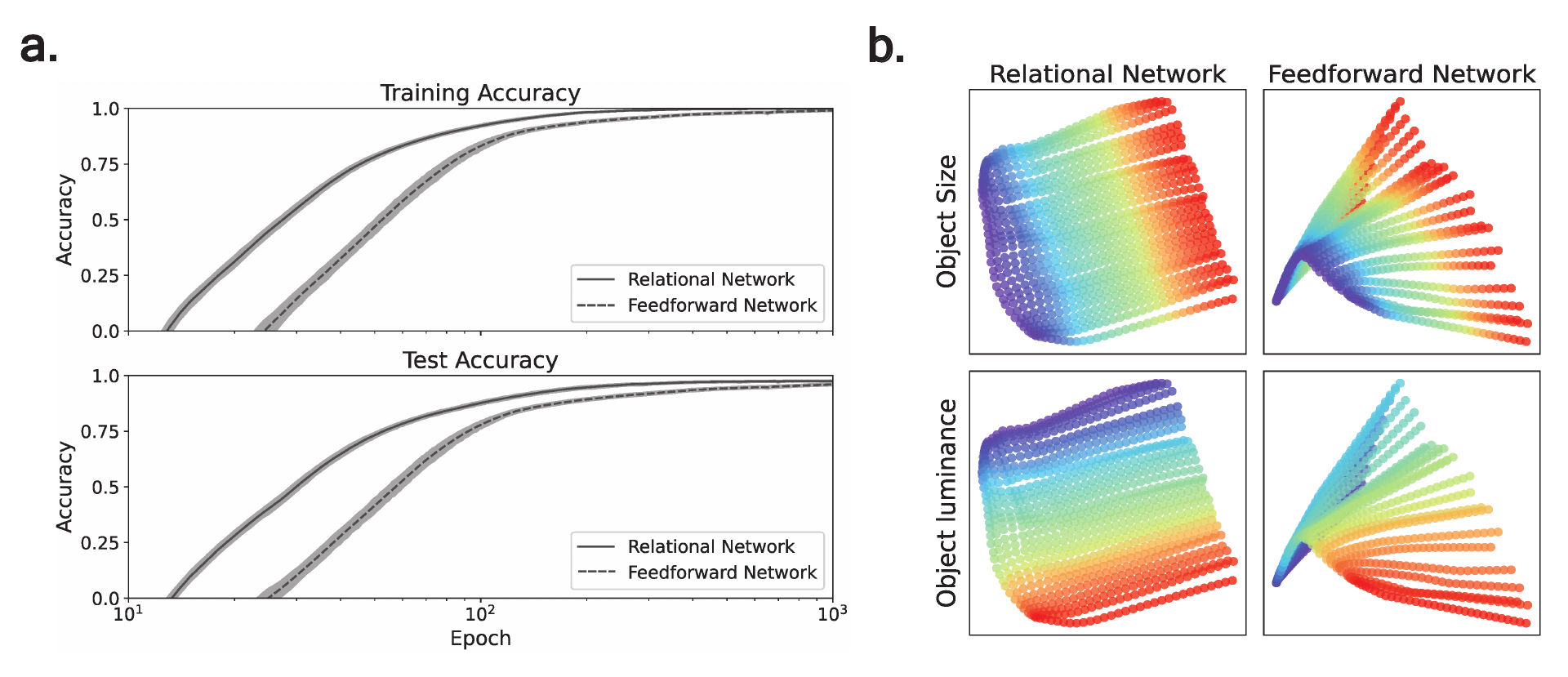}
\caption{\textbf{(a)} Learning curves for training loss and generalization performance indicate that the relational architecture learns more rapidly than the feedforward architecture. \textbf{(b)} 2-dimension PCA of network embeddings learned by each network. Note that relational network learns orthogonal representations for each dimension, whereas the feed-forward network learns a non-linear manifold.}
\label{fig:exp1_results}
\end{figure*}

\section{Dimensional Representations Align with Human Behavior}

Here, we examined the extent to which the relational bottleneck can explain findings in human performance reflecting biases towards regularity (e.g., symmetry and parallelism) in geometric figures, and that have previously been hypothesized to reflect the presence of pre-specified symbolic representations \citep{sable2021sensitivity}. To do so, we trained the network described above (Figure 1a, red pathway) on an oddball detection task involving simple quadrilateral figures, used in \cite{sable2021sensitivity} to compare the performance of humans to non-human primates and artificial agents. That study was offered as evidence not only that humans exhibit a regularity bias in processing geometric figures that is not observed in non-human primates, but also that this bias can be captured by models that are explicitly imbued with pre-specified symbolic primitives but not standard deep learning neural networks that are trained directly on the images. Here, we tested whether a network imbued with a relational bottleneck — but not any symbolic primitives — could reproduce the regularity bias observed in humans.

\subsection{Methods}
The relational network described above (Fig. 1, red pathway) and SimCLR — a standard contrastive learning method, implemented here using a ResNet encoder — were trained and tested on approximately $60,000$ trials using the same stimuli and following the same protocol used with humans and non-human primates in \cite{sable2021sensitivity}. On each trial during the test phase, each network was presented with six images of quadrilateral figures, five of which were symmetric and identical in shape but varied in size and/or rotation, and one of which — the “oddball” — was different in shape (Fig. 3). The oddballs were constructed by perturbing the bottom right vertex of the reference shape to violate its regularity in the same way as in  \cite{sable2021sensitivity}. For each trial, we extracted the embedding in the model for each of the six images, and used these to identify the oddball as the one that was furthest from the centroid of the set (defined as the mean of the embeddings). We then computed the average error rate for each shape category and correlated the model error rates with the human and non-human primate error rates.

\begin{figure}[h]
\centering 
\includegraphics[width=\linewidth]{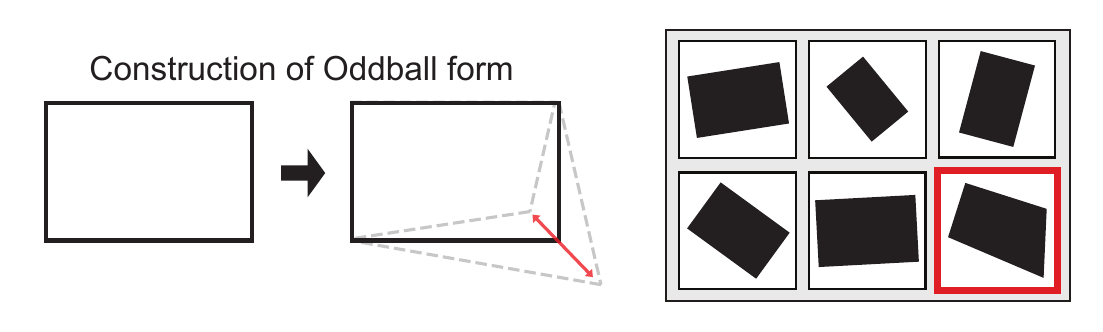}
\caption{\textbf{Oddball construction \& trial structure} Stimuli consisted of quadrilateral forms varying in their regularity/symmetry. Each trial was comprised of five variants of the same stimulus varying only in size and rotation, and one "oddball" stimulus constructed by perturbing the bottom right vertex of the reference stimulus to violate its regularity. Participants and networks were evaluated on their accuracy in identifying the oddballs.}
\label{fig:exp2_methods}
\end{figure}

\begin{figure*}[h!]
\centering 
\includegraphics[width=450px]{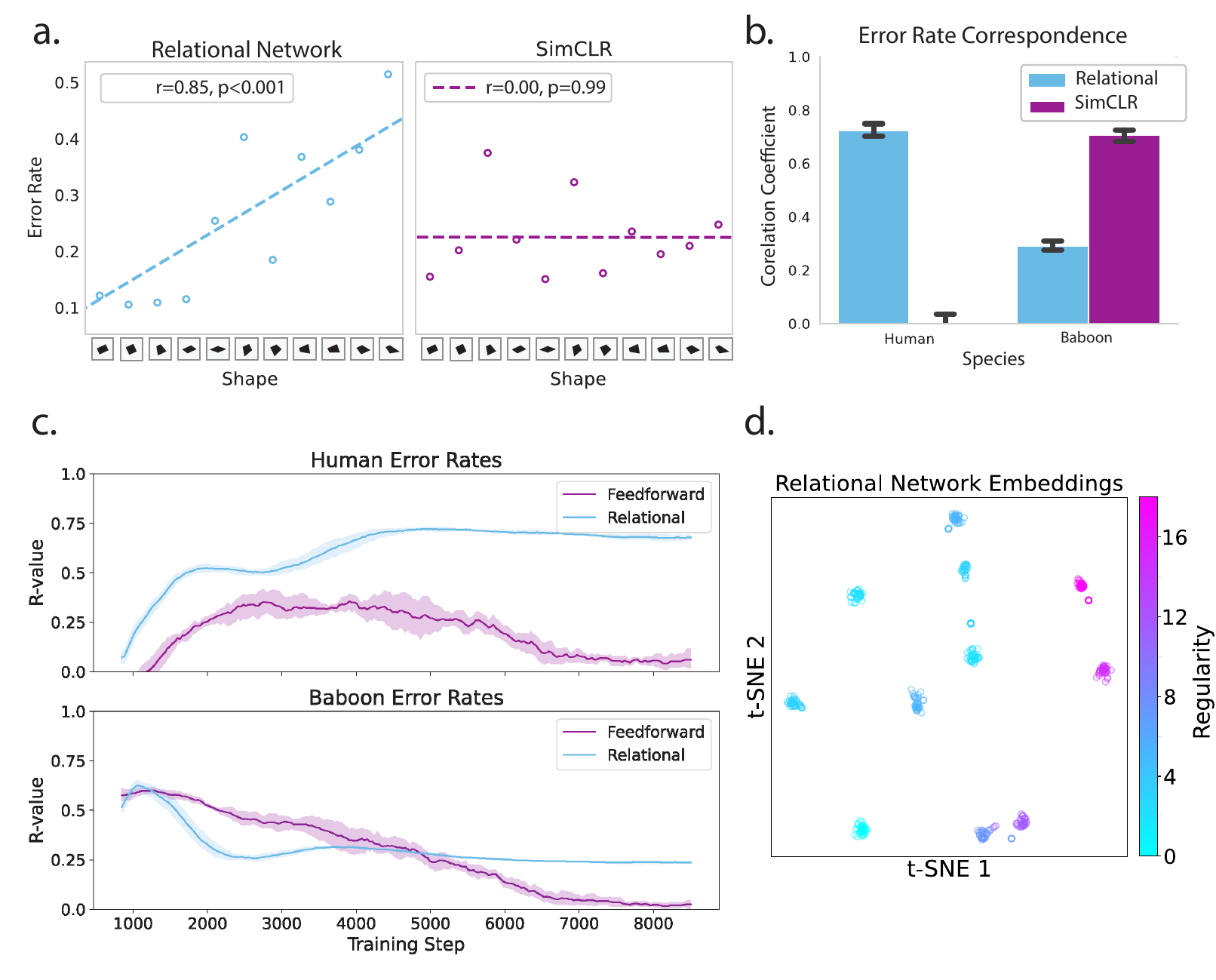}
\caption{\textbf{Relational Network and SimCLR performance on oddball task:} \textbf{(a)} Error rates for the relational network and SimCLR at representative points during training. Note that the relational network's error rates exhibit a positive slope as a function of decreasing geometric regularity, consistent with human performance on this task, while the SimCLR network displays no sensitivity to geometric regularity. \textbf{(b)} Correlation coefficients between the model error rates and human and baboon error rates. Note that SimCLR most closely resembles baboon performance on this task while the relational network most closely corresponds with the human error rates. \textbf{(c)} t-SNE plot of reference shape embeddings from the relational network colored according to stimulus regularity calculated as the sum of the binary symbolic properties for each stimulus.}
\label{fig:exp2_results}
\end{figure*}

\subsection{Results}
We found that the network exhibited a pattern of performance as a function of both learning and regularity of shapes that closely resembled that observed by \cite{sable2021sensitivity} for humans but not for non-human primates (Fig. 4a-b). The network exhibited this pattern despite the fact that it was not endowed with any pre-specified symbolic primitives, nor any specific inputs that would have made these easier to discover. Rather, the results are attributable to the presence of the relational bottleneck, which has been shown in other settings to predispose to the discovery of low dimensional, abstract representations that can function as symbols \citep{webb2020emergent, kerg2022neural, altabaa2023abstractors}. Moreover, these biases towards human-like performance on this task emerge early during learning and are persistent over the course of network training (Fig. 4c). The relational training scheme yielded networks that  learned well structured representations of the underlying category structure of the task (Fig 4d). While standard contrastive training has also been shown to produce networks sensitive to category information, our objective was to instead evaluate the degree of representational factorization within both relational and standard contrastive network architectures. 

To do so, we assessed whether features associated with stimulus regularity were distinctly encoded in the network embeddings by applying linear regression models on the first 50 principal components of the networks' embeddings using 20-fold cross validation. The results indicated a more robust representation of stimulus regularity in the relational network ($R^2=0.89$) compared to the standard contrastive model ($R^2=0.68$). Notably, these findings were observed despite the fact that classifiers more accurately predicted the stimulus category from the standard contrastive networks' representations compared to the relational network representations ($92\%$ and $86\%$, respectively). This suggests that the improved decoding of stimulus regularity in the relational network is not due to a clearer representation of stimulus category, and is instead a product of the network's tendency to encode features along implicitly factorized and more linearly separable subspaces within the relational networks' embeddings.

\section{Discussion}
Low dimensional, factorized representations that can be used compositionally are generally thought to be a foundational requirement on which the flexibility of computation exhibited by humans is built. Most models that seek to describe such forms of computation have either used traditional symbolic processing mechanisms \citep{anderson2013adaptive, newell1994unified} or have explicitly imbued neural networks with pre-specified symbolic primitives over which they can operate \citep{garcez2002neural}. Some neural network models have used unsupervised learning coupled with specialized loss functions \citep{higgins2016beta, kim2018disentangling}, but these require sensitive hyper-parameter tuning to discover dimensional structure. Another approach has been to supply the latent factors as fully supervised training targets with generative models \citep{tran2017disentangled}. Our method provides an intermediate form of supervision between unsupervised and fully supervised methods. By using a scalar similarity target that implicitly captures the task’s relational structure along the relevant dimensions, it implements a simple relational bottleneck \citep{webb2023relational} that, in turn, encourages the emergence of factorized representations, without explicitly providing them as targets. This approach affords the network the benefits of encoding task relevant features in its embeddings, while also remaining flexible enough to encode other features that may be relevant to the performance of other downstream tasks. 

This method may also better account for how children construct factorized representations of their environments by receiving signals about the similarity structure of the world, either from cues in the environment or from explicit instruction by teachers and/or parents \citep{markman1984children, gelman1986categories}. Given the constancy of objects implicit in real-world environments, viewing these objects under multiple viewing conditions and in contrast to distinct objects may help carve out representations that factorize the relevant dimensions of variation. Furthermore, explicit instruction about the similarity of novel objects compared to known ones may implicitly provide information about the dimensional structure of the world that further facilitates factorization that can be exploited by architectures that include a form of relational bottleneck. In this way, agents may learn factorized representations of latent features without requiring explicit instruction about all of the particular features of the relevant stimuli. Instead, a training signal providing rich information about the similarity of objects in combination with the appropriate mechanisms for computing relations may be sufficient to leverage this comparatively weak form of supervision to learn factorized representations.

Furthermore, if an agent’s experience does not encompass a sufficiently rich sampling of the underlying feature space, as is often the case in many naturalistic learning settings, relational architectures may insulate agents from the risks of overfitting that are common in more traditional architectures \citep{srivastava2014dropout, goodfellow2016deep}. This protective effect is especially pronounced in environments where latent features are categorical, and stimuli consist of various combinations of these categorical features (see Supplementary information for a more detailed consideration and examples of this point). In cases of sparse feature sampling, traditional neural networks tend to overfit by picking up on spurious correlations across independent feature dimensions. Relational bottleneck models, however, are more resistant to this, thereby reducing the likelihood of overfitting in such situations \citealp{webb2023relational}. Moreover, the benefits of relational processing in promoting generalization and rapid learning are not only present in simple representation learning tasks as tested in the simulations reported here, but also across a range of challenging visual reasoning and analogy tasks \citep{mondal2023learning, webb2020emergent}. 

Previous work has demonstrated how augmenting neural networks with an explicit mechanism for computing relations provides substantial benefits in learning efficiency and generalization in navigational tasks \citep{whittington2020tolman} and in basic reasoning tasks \citep{altabaa2023abstractors, webb2020emergent}, approaching the sample efficiency and generalization abilities of human learners in these domains. Here, we show that these same computational elements may favor the formation of factorized upstream representations, that facilitate the discovery of relational structure.

This architecture may also provide structural/mechanistic insights into human brain function. Several studies have indicated that regions of the medial temporal lobe and hippocampus play an important role in navigation by providing a mechanism for binding features \citep{whittington2020tolman} and computing the similarity of the current state with those stored in memory \citep{norman2003modeling, o1994hippocampal}. Such machinery for estimating similarity relations among distinct representations may provide a powerful architectural inductive bias not just useful memory retrieval, but also for factorizing representations. This hypothesis is consistent with recent evidence suggesting that regions that provide direct input to the hippocampus such as the entorhinal cortex encode highly factorized representations of space, time, and other cognitive variables \citep{fyhn2004spatial, aronov2017mapping, constantinescu2016organizing, chandra2023high}. The work reported here provides one account for why these representations may emerge in regions providing input to the hippocampus. Furthermore, the framework suggests that other structures and mechanisms in the brain, that have similar functional attributes, may support relational abstraction and representational factorization in other domains (for example, the cerebellum and parietal cortex in the domain of motor function \citep{ravizza2006cerebellar, d2020evidence, mcdougle2022continuous}).

\section{Conclusion}
Our findings extend previous work, which has demonstrated that the use of a relational bottleneck \citep{webb2023relational} can induce a network to learn abstract rules and use these for extreme forms of generalization \citep{webb2020emergent, kerg2022neural, altabaa2023abstractors}. We show that the same inductive bias can induce the system to discover factorized, compositional representations of feature dimensions relevant to task performance in a data efficient manner, and in a form that approximates the efficiency of coding and flexibility of processing exhibited by the human brain.

\bibliographystyle{apacite}

\setlength{\bibleftmargin}{.125in}
\setlength{\bibindent}{-\bibleftmargin}

\bibliography{CogSci_Template}

\clearpage % Ensure we start on a new page
\onecolumn
\section{Supplementary Information}

\begin{figure*}[h]
\centering 
\includegraphics[width=400px]{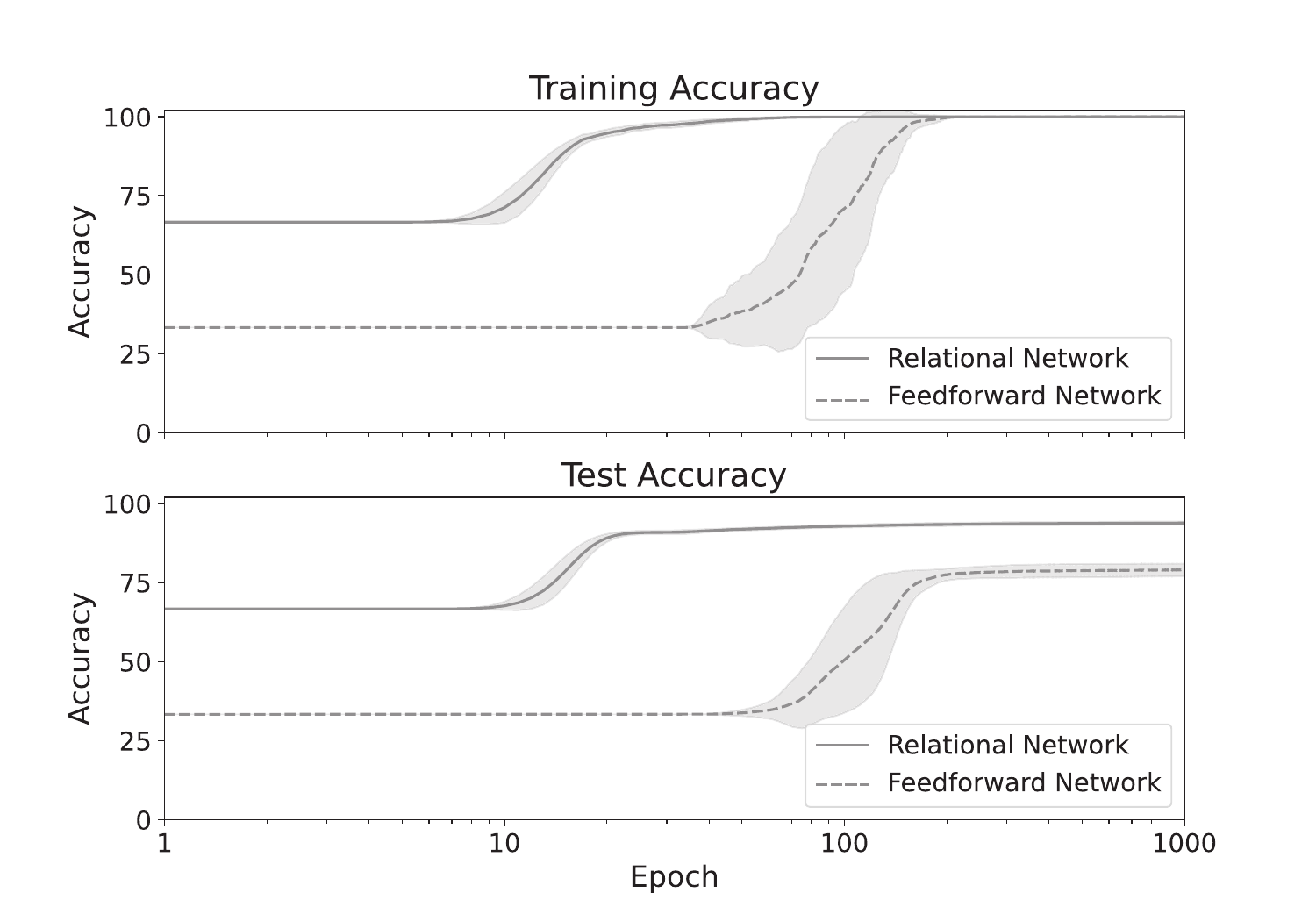}
\caption{\textbf{Performance on category task} We trained ten relational and feedforward neural networks, which differed solely in their initial weight configurations, on a categorical same/different task. The task involved simple binary stimuli, each represented by two one-hot encoded features. Each feature could take one of 30 possible values, resulting in a total of 900 unique stimuli. For training, we randomly selected a training set of 30 stimuli, which constituted $3.\overline{3}\%$ of the entire stimulus set. The objective for the networks was to learn to make similarity judgments on this holdout set. During the testing phase, we evaluated the networks' performance on the remaining $96.\overline{6}\%$ of the stimuli, which they had not seen during training. The networks' were evaluated on their ability to accurately judge the similarity of previously unseen stimuli. Performance on the task was computed as accuracy on the task after binarizing the networks' similarity judgements by thresholding. Both the relational and feedforward networks achieved perfect accuracy on the training set. Notably, the relational network learned to perform the task more rapidly than the feedforward network. In terms of generalization, the relational network's performance on the holdout set was nearly perfect, while the feedforward network's performance was significantly lower. Accuracy on the task is higher initially in the relational network due to the relational bottlenecks sensitivity to the trivially separated features present in the input data.}
\label{fig:exp2_methods}
\end{figure*}

\end{document}